\documentclass{article}

\usepackage{iclr2018_conference,times}
\usepackage{url}

\usepackage{amsmath}
\usepackage{amsfonts}
\usepackage{amssymb}
\usepackage{mathtools}
\usepackage{graphicx}
\usepackage{hyperref}
\usepackage{subcaption}

\newcommand{\squeeze}{\vspace{-2.5mm}}

\title{On the Dimensionality of Embeddings \\ for Sparse Features and Data}

\author{Maxim Naumov \\ Facebook, 1 Hacker Way, Menlo Park, CA, 94025}

\begin{document}

\maketitle

\begin{abstract}
In this note we discuss a common misconception, namely that embeddings are always used to reduce the dimensionality of the item space. We show that when we measure dimensionality in terms of information entropy then the embedding of sparse probability distributions, that can be used to represent sparse features or data, may or not reduce the dimensionality of the item space. However, the embeddings do provide a different and often more meaningful representation of the items for a particular task at hand. Also, we give upper bounds and more precise guidelines for choosing the embedding dimension.
\end{abstract}

\section{Background}

The mapping of concepts, objects or items into a vector space is called an embedding. The representation of $n$ items in $d$ dimensional vector space is used in many applications, which can broadly be split into two categories. 

The first category is characterized by models that produce embeddings. In this case the embeddings represent the information obtained by the model about the item space. If the model represents a probability distribution over items in the dataset then we can interpret the output as an embedding of this probability distribution.

For instance, singular value decomposition \cite{Golub2012} is used in latent semantic analysis/indexing (LSA/LSI) \cite{Dumais2005} to produce low-rank approximations of the original data. In this case a low-rank approximation of the word-document matrix $C \in \mathbb{R}^{m \times n}$ for $m$ words and $n$ documents, where each row corresponds to a word and each column corresponds to a document, while non-zero entries represent the occurence of words in documents, can be written as 
\begin{equation}
C \approx WV^T
\label{eq:lowrank}
\end{equation}
where the matrices $W \in \mathbb{R}^{m \times d}$ and $V \in \mathbb{R}^{n \times d}$ can be interpreted as embeddings of words and documents in a $d$ dimensional space. A similar interpretation can be given to matrix factorization for collaborative filtering, with the caveat that empty matrix entires are unknown (rather than being $0.0$) \cite{koren2009matrix,frolov2017tensor}. 

Also, deep learning models based on auto-encoders \cite{Sedhain2015}, multi-label classification \cite{Bengio2010} and neural machine translation (NMT) \cite{Neubig2017} can be seen as generating a probability distribution over a set of classes/objects. Let this probability distribution be represented as a vector $\textbf{p} \in \mathbb{R}^{n}$. Notice that the embeddings produced by the models, such as embeddings of target language words in NMT, represent an embedding of this probability distribution over $n$ items into $d$ dimensional space
\begin{equation}
\textbf{v} =  V^T \textbf{p} 
\label{eq:probemb}
\end{equation}
for some embedding matrix $V \in \mathbb{R}^{n \times d}$. Notice that during training we often start with an arbitrary probability distribution and therefore vector $\textbf{p}$ is dense. However, when training converges towards the end it is common for the probability distribution to concentrate more at particular items and therefore vector $\textbf{p}$ becomes sparse (and if needed it can be quantized to eliminate a long tail of small values). It will become clear in this note that the information the embedding $V$ needs to represent will vary greatly based on the signature of vector $\textbf{p}$. In particular, when $\textbf{p}$ is very sparse and its elements can be quantized to a few values the embedding dimension $d$ needed to represent the information in it can be relatively small.

The second category is characterized by models that consume embeddings. In this case embeddings represent information obtained from the input features or data. It is common for these input features to be sparse, such as the history of user clicks on webpages or words present in a post. The sparse features are often represented by a list of integer indices that select items from a larger sequence/set (in contrast to dense features that are represented by a single floating point number). They can also be represented by a sparse vector $\textbf{p}$ as will be shown in the next section.  

For instance, neural collaborative filtering \cite{he2017neural}, wide \& deep \cite{cheng2016wide} and deep \& cross recommendation systems \cite{wang2017deep} all use embeddings to process sparse input features. The advantage of using embeddings instead of lists of sparse items is that we can measure distance between them in a more meaningful way. Also, notice that embedding elements represent the sparse features in some abtract space relevant to the model at hand, while integers simply represent an ordering of the input data.

The natural language processing models \cite{kalchbrenner2013recurrent, sutskever2014sequence} may fall somewhere in between these two categories. In particular, NMT models \cite{Neubig2017} often use two embeddings one representing words in the source and another in the target language. On one hand, the source embedding can be seen as a sparse feature consumed by the model. It is characterized by a list of indices that selects words used in the input sentence. On the other hand, the target embedding can be seen as a representation of the probability distribution over words in the target language.   

In this note we focus on the embedding of sparse vectors $\textbf{p}$ that can be used to represent sparse features and data that belong to the second category. We point out that the implications of choosing a particular value of a hyper-parameter $d$ are not theoretically well understood. The choice is usually based on empirical experiments or resource limitations, such as compute and memory available on hardware platforms \cite{Jongsoo2018}. A recent work attempted to explain the sizing of the embedding vectors based on pairwise inner product dissimilarity metric \cite{yin2018}. 

We propose an alternative approach based on the entropy information measure. We leverage the ideas of \cite{Naftali1999,ShwartzZiv2017} as well as \cite{Traub1980,Pinkus1985,Donoho2006}, but in contrast to them we do not attempt to explain the behavior of neural networks or find ways to compress the parameters/data. We use the information measure to discuss the dimensionality and provide guidelines on sizing of the embedding vectors, i.e. we provide roofline and more precise models for selecting the dimensionality $d$.  

\section{Introduction to embeddings}

Let $n$ items be mapped into a $d$ dimensional vector space. The vectors corresponding to $n$ items are often organized into an embedding table, which can be seen as a tall matrix $V \in \mathbb{R}^{n \times d}$ with $n \gg d$  and that can be written as 
\begin{equation}
V^T = [\textbf{v}_1, ..., \textbf{v}_n]
\label{eq:embedding}
\end{equation}
where vector $\textbf{v}_i$ corresponds to $i$-th item.

A sparse feature is characterized by a list of integer indices, which can be represented as item lookups with different signature in the embedding table.

The item lookup with a single index is often encoded as a dense matrix-vector multiplication
\begin{equation}
\textbf{v}_i = V^T \textbf{e}_i 
\label{eq:itemlookup}
\end{equation} 
where vector $\textbf{e}_i \in \mathbb{R}^{n}$ and
\begin{equation}
\textbf{e}_i^T = [0,...,1,...,0]
\label{eq:onehotvector}
\end{equation}
has $1$ in $i$-th position and $0$ everywhere else. It is often referred to as a one-hot encoding vector.

Notice that we can can select multiple items with some weights in a single lookup and express it as
\begin{equation}
\textbf{u} = V^T \textbf{a} = a_{i_1} \textbf{v}_{i_1} + ... + a_{i_k} \textbf{v}_{i_k}
\label{eq:multipleitemlookup}
\end{equation}
where vector $\textbf{a} \in \mathbb{R}^{n}$ and
\begin{equation}
\textbf{a}^T = [0,...,a_{i_1},...,a_{i_k},...,0]
\label{eq:multiplehotvector}
\end{equation}
has weight $a_{i} \ne 0$ at $i=i_1,...i_k$ and $0$ everywhere else \cite{jia2014caffe,PyTorch}. 

Further, we can generalize it to multiple lookups where each selects multiple items with some weight and encode it as a dense matrix-sparse matrix multiplication
\begin{equation}
U^T = V^T A
\label{eq:matrixlookup}
\end{equation}
where sparse matrix $A \in \mathbb{R}^{n \times r}$ and
\begin{equation}
A = [\textbf{a}_1,...,\textbf{a}_r]
\label{eq:matrixhotvectors}
\end{equation}
is composed of multiple vectors $\textbf{a}_j$ each corresponding to a single lookup\footnote{Notice that the vector subscript in here has a different meaning than before. In \eqref{eq:matrixhotvectors} it denotes the $j$-th lookup, while in in \eqref{eq:onehotvector} it denoted the $j$-th item being selected.} with non-zero elements corresponding to the items being selected. The output matrix $U \in \mathbb{R}^{r \times d}$ is the result of $r$ lookups.

This setup is often used to state that the embedding vectors project $n$ dimensional items space into $d$ dimensional embedding vectors. In the next section we will examine this statement in more detail and show that it can be misleading and therefore lead to incorrect conclusions.  

\section{The Dimensionality of embeddings}

Notice that when we discussed space dimensions in the previous section, we never took into consideration the precise vector element type and how much information we can represent with it. Let us now incorporate it into our analysis by measuring the cardinality of the set it can describe and the information associated with it.

Recall that the entropy of an information source $s$ is given by 
\begin{equation}
H(s) = - \sum_{i=1}^n p_i \log_2 p_i
\end{equation}
where $p_i$ is the probability of $i$-th symbol to be communicated \cite{Shannon1949}.

Notice that the embedding vector has $d$ elements, each with $s$ bits. Therefore, it could represent
\begin{equation}
g = 2^{ds}
\end{equation}
values and if we interpret it as an information source $\textbf{v}_{\_}$ its entropy is 
\begin{equation}
H(\textbf{v}_{\_}) = - \sum_{i=1}^{g} p_i \log_2 p_i
\label{eq:embeddingentropy}
\end{equation}
where $p_i$ denotes the probability of $i$-th value being selected. Hence, if $p_i = 1/g$ is uniform then
\begin{equation}
H(\textbf{v}_{\_}) = ds
\label{eq:embeddingentropy2}
\end{equation}

\subsection{Single lookup of single item}

Let a single lookup of a single item be done as shown in \eqref{eq:itemlookup}. In this case the $i$-th embedding vector represents the $i$-th item from the item space. 

Notice that because we represent this lookup as a binary vector $\textbf{e}_i$ in \eqref{eq:onehotvector} in $n$ dimensional space, it can describe only $n$ items and if we interpret it as an information source $\textbf{e}_{\_}$ its entropy is
\begin{equation}
H(\textbf{e}_{\_}) = - \sum_{i=1}^n p_i \log_2 p_i
\label{eq:onehotentropy}
\end{equation}
where $p_i$ denotes the probability of $i$-th item being selected. Note that if $p_i = 1/n$ is uniform then
\begin{equation}
H(\textbf{e}_{\_}) = \log_2 n
\label{eq:onehotentropy2}
\end{equation}

Therefore, the dimensionality of the item and embedding spaces, as measured by the information they can represent, can be compared by looking at \eqref{eq:embeddingentropy} and \eqref{eq:onehotentropy}. For instance, under the assumption of uniform probability of item/value selection, if $n=20$M and using \eqref{eq:onehotentropy2} we have $H(\textbf{e}_{\_}) \approx 24.3$ then a single 32-bit element is enough to represent information in the item space.

\subsection{Single lookup of multiple items}
Let a single lookup of multiple items be done as shown in \eqref{eq:multipleitemlookup}. In this case the $i$-th embedding vector represents combinations of $i$-th and other items from the item space.

Let us first consider the case when this lookup is a binary vector in $n$ dimensional space, i.e. $a_i \in \{0,1\}$ for $i=1,...,k$ in \eqref{eq:multiplehotvector}. In this case, the vector can describe $h$ items, with
\begin{equation}
h = \binom{n}{k} = \frac{n!}{k!(n-k)!}
\end{equation}
and if we interpret it as an information source its entropy is
\begin{equation}
H(\textbf{a}_{\_}) = - \sum_{i=1}^h p_i \log_2 p_i
\label{eq:multiplehotentropy}
\end{equation}
where $p_i$ is the probability of a combination being selected. Note that if $p_i = 1/h$ is uniform then
\begin{eqnarray}
H(\textbf{a}_{\_}) &=& \log_2 h \label{eq:multiplehotentropy_exact} \\ 
             &\approx& n \log_2 \frac{n}{n-k} + k \log_2 \frac{n-k}{k} + \label{eq:multiplehotentropy_ramanujan} \\ \nonumber
                    && \phantom{1} + \frac{1}{6} \log_2 \frac{n+4n^2+8n^3}{((n-k)+4(n-k)^2+8(n-k)^3)(k+4k^2+8k^3)} - \frac{1}{2} \log_2 \pi 
\end{eqnarray}
where we have used properties of logarithms and Ramanujan's approximation \cite{Ramanujan1988}
\begin{equation}
\ln n! \approx n \ln n - n + \frac{1}{6} \ln (n+4n^2+8n^3) + \frac{1}{2} \ln \pi
\label{eq:ramanujan}
\end{equation}
that are described in more detail in the appendix.

Therefore, the dimensionality of the item and embedding spaces, as measured by the information they can represent, can be compared by looking at \eqref{eq:embeddingentropy} and \eqref{eq:multiplehotentropy}. For instance, under the assumption of uniform probability of item/value selection, if $n=20$M, $k=100$ and using \eqref{eq:multiplehotentropy_ramanujan} we have $H(\textbf{a}_{\_}) \approx 1756.3$ then a vector with 64 elements with 32-bit per element is enough to represent information in the item space.

\subsection{Single lookup of multiple items with weights}

Let us now consider the case when this lookup is a vector in $n$ dimensional space, with each of the $a_i$ represented by $t$ bits for $i=1,...,k$ in \eqref{eq:multiplehotvector}. Notice that in this case the analysis of the previous section remains the same, but we can now select $2^t$ values for each position in the lookup. Therefore, the vector can describe $h'$ items, with
\begin{equation}
h' = 2^{tk} h
\label{eq:multiplehotentropy_weigths}
\end{equation}
Note that if we interpret it as an information source, under the assumption $p_i = 1/h'$ is uniform, then its entropy is
\begin{eqnarray}
H(\textbf{a}_{\_}) &=& \log_2 h' = tk + \log_2 h \\
             &\approx& tk + n \log_2 \frac{n}{n-k} + k \log_2 \frac{n-k}{k} + \label{eq:multiplehotentropy_ramanujan2} \\ \nonumber
                    && \phantom{12345} + \frac{1}{6} \log_2 \frac{n+4n^2+8n^3}{((n-k)+4(n-k)^2+8(n-k)^3)(k+4k^2+8k^3)} - \frac{1}{2} \log_2 \pi
\end{eqnarray}

Therefore, the dimensionality of the item and embedding spaces, as measured by the information they can represent, can be compared by looking at \eqref{eq:embeddingentropy} and \eqref{eq:multiplehotentropy_weigths}. For instance, under the assumption of equal probability of item/value selection, if $n=20$M, $k=100$, $t=16$ and using \eqref{eq:multiplehotentropy_ramanujan2} we have $H(\textbf{a}_{\_}) \approx 3356.3$ then a vector with 128 elements with 32-bit per element is enough to represent information in the item space.

\subsection{The effects of mini-batch}

We point out that the use of mini-batch of size $r$ does not affect the dimensionality because each vector in the mini-batch is treated independently. This can be observed in \eqref{eq:matrixlookup} where $r$ lookups from matrix $A$ in embedding $V$ generate $r$ results in matrix $U$.  

\section{Upper bounds and sizing guidelines for embeddings}

Notice that embeddings corresponding to sparse features do not necessarily reduce the dimensionality of data, as measured by the information they can represent. The dimensionality reduction depends on the size of the embedding vectors and the detailed signature of the input lookup vectors. 

The upper bound (roofline) dimensionality of different types of lookups is provided in \eqref{eq:onehotentropy2}, \eqref{eq:multiplehotentropy_ramanujan} and \eqref{eq:multiplehotentropy_ramanujan2} which can be compared with dimensionality of embeddings given in \eqref{eq:embeddingentropy2}. We have tabulated a few sample lookup signatures and embedding dimensions for comparision in Tab. \ref{tab:dimension_roofline}. 

\begin{table}[h]
\centering
\begin{tabular}{l|l|c|c|c|c|c|c|}
\multicolumn{3}{c|}{Input Lookup Signature} & \multicolumn{4}{c}{Embedding Dimensions} \\
\hline             
n   &  k   & H(.)   & d (s=8)  & d (s=16) & d (s=32) & H(.) \\
\hline
1M    & 1       & 19.9     & 4       & 2       &  1      & 32      \\
10M   & 1       & 23.2     & 4       & 2       &  1      & 32      \\
100M  & 1       & 26.5     & 4       & 2       &  1      & 32      \\
1M    & 10      & 163.0    & 24      & 12      &  6      & 192     \\
10M   & 10      & 196.3    & 28      & 14      &  7      & 224     \\
100M  & 10      & 229.5    & 32      & 16      &  8      & 256     \\
1M    & 100     & 1324.1   & 168     & 84      &  42     & 1344    \\
10M   & 100     & 1656.3   & 208     & 104     &  52     & 1664    \\
100M  & 100     & 1988.5   & 252     & 126     &  63     & 2016    \\
1M    & 1000    & 9958.0   & 1248    & 624     &  312    & 9984    \\
10M   & 1000    & 13281.2  & 1664    & 832     &  416    & 13312   \\
100M  & 1000    & 16603.3  & 2076    & 1038    &  519    & 16608   \\
\end{tabular}
\caption{Entropy of a sample set of lookup signatures and embedding dimensions}
\label{tab:dimension_roofline}
\end{table}

\begin{figure}[h]
 \begin{center}
  \begin{subfigure}[b]{0.49\textwidth}
   \includegraphics[width=\textwidth]{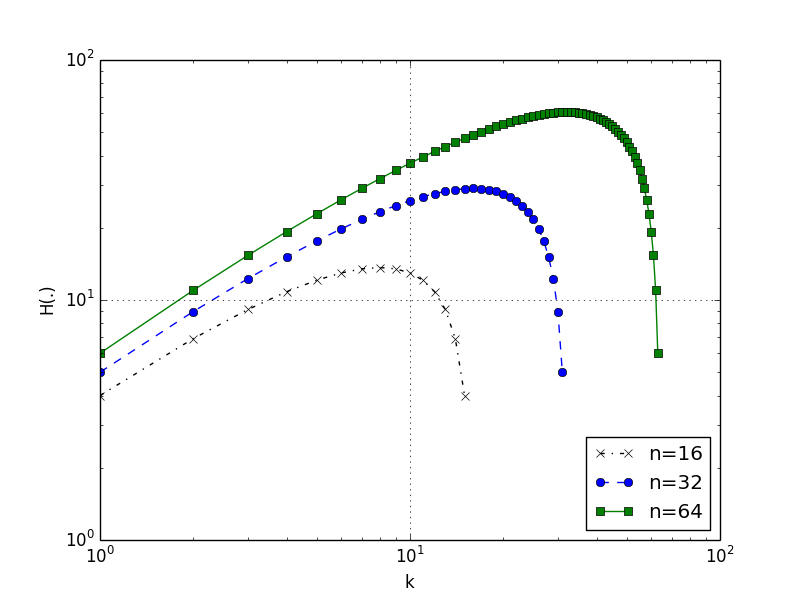}
   \caption{Entropy \eqref{eq:multiplehotentropy_exact} for $n=64$ and $k=1,...,63$}
   \label{fig:entropy_full}
  \end{subfigure}
  \begin{subfigure}[b]{0.49\textwidth} 
   \includegraphics[width=\textwidth]{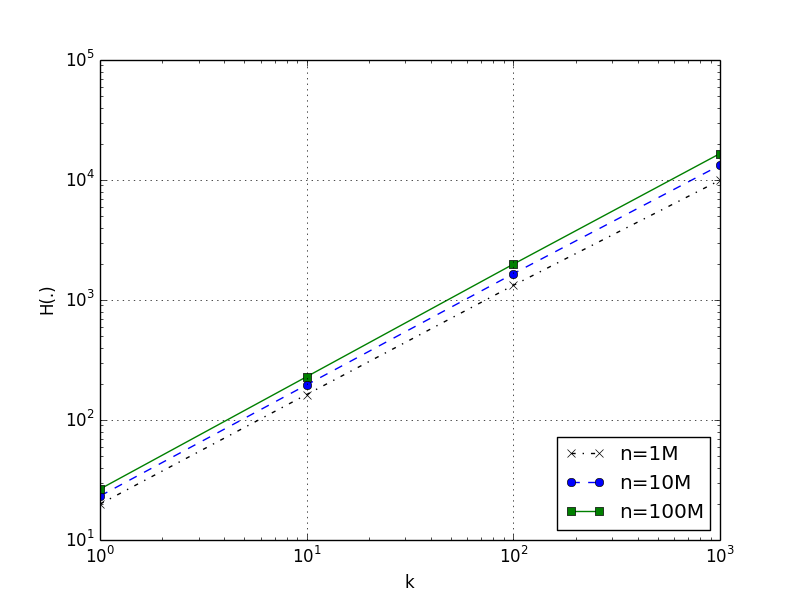}
   \caption{Entropy \eqref{eq:multiplehotentropy_exact} for sample lookup signatures}
   \label{fig:entropy_zoom}
  \end{subfigure} 
  \caption{Entropy of a sample set of lookup signatures in log scale}
  \squeeze \squeeze
 \end{center}
\end{figure}

Although, the function \eqref{eq:multiplehotentropy_exact} achieves its maximum for larger $n$ as shown in Fig. \ref{fig:entropy_full}, notice that for the sample lookup signatures in Tab. \ref{tab:dimension_roofline} the choice of $k$ has a much larger effect on the entropy function, as shown on Fig. \ref{fig:entropy_zoom}. We point out that the function is ploted on a log scale in both plots.

In practice not all combinations will be exercised by the item lookups. We can make a pass over the input dataset to discover how many combinations are actually present and how often the same combinations are repeated. Then, we can use formulas \eqref{eq:onehotentropy} and \eqref{eq:multiplehotentropy} to estimate the dimensionality of the input dataset from the information perspective. Ultimately, the more precise measure of dimensionality of the input dataset can guide a better choice of the embedding vector size and elemet type to be used in the model. 

Once again we point out that although embeddings do not necessarily reduce the dimensionality of lookups, they do provide a different and very useful representation of them. Note that the embedding values are learned during training. Therefore, the embeddings are found based on what the lookups represent/mean in some abtract space relevant to the model at hand. 

\section{Conclusion}

We have discussed embeddings corresponding to dense and sparse probability distributions. We have analyzed the dimensionality of the item lookups and embeddings corresponding to sparse features and data. We have shown that using embeddings does not necessarily reduce the dimension of the lookups, as measured in terms of the information they can represent. We have also provided rooflines and more precise guidelines for choosing the embedding size for the dataset and model at hand.

\section*{Acknowledgements}

The author would like to thank Aleksandr Ulanov, Dheevatsa Mudigere, Satish Nadathur and Misha Smelyanskiy for thoughtful comments as well as Mark Tygert and Juan Miguel Pino for insighful discussion on the use of embeddings in different applications and NMT models.

\bibliography{embedding_dimensionality_reduction}
\bibliographystyle{iclr2018_conference}

\clearpage
\newpage

\section{Appendix}



\subsection{Logarithmic Indetities}

We list a few handy logarithmic identities in this section \cite{Cormen2009}. The multiplication and division operations are equivalent to
\begin{equation}
\log_b xy = \log_b x + \log_b y
\end{equation}
and
\begin{equation}
\log_b x/y = \log_b x - \log_b y
\end{equation}
respectively. Also, we can change logarithm from base $b$ to $c$ using the following 
\begin{equation}
\log_b x = \log_c x / \log_c b
\end{equation}

\subsection{Approximation theory}

There exist a number of approximations to the factorial function, including Stirling's approximation \cite{LeCam1986,Romik2000}
\begin{equation}
n! \sim \sqrt{2 \pi n} \left(\frac{n}{e}\right)^n 
\end{equation}
and more accurate Ramanujan's approximation \cite{Ramanujan1988,Karatsuba2001}
\begin{equation}
n! \sim \sqrt{2 \pi n} \left(\frac{n}{e}\right)^n \left( 1 + \frac{1}{2n} + \frac{1}{8n^2} \right)^{1/6}
\label{eq:ramanujan_fact_approx}
\end{equation}

Then, using \eqref{eq:ramanujan_fact_approx} it follows that
\begin{equation}
\ln n! \sim n \ln n - n + \frac{1}{6} \ln (n+4n^2+8n^3) + \frac{1}{2} \ln \pi
\label{eq:ramanujan_logfact_approx}
\end{equation}

Finally, notice that using logarithmic properties and \eqref{eq:ramanujan_logfact_approx} we can write
\begin{eqnarray}
\log_2 \frac{n!}{k!(n-k)!} &=& \log_2 n! - \log_2 (n-k)! - \log_2 k! \nonumber \\
        &\approx& \phantom{+}\frac{1}{\ln 2} \left(n \ln n - n + \frac{1}{6} \ln (n + 4n^2 +8n^3) + \frac{1}{2} \ln \pi \right)     \nonumber \\
                &&         - \frac{1}{\ln 2} \left((n-k) \ln (n-k) - (n-k) + \frac{1}{6} \ln ((n-k) + 4(n-k)^2 +8(n-k)^3) + \frac{1}{2} \ln \pi \right) \nonumber \\
                &&         - \frac{1}{\ln 2} \left(k \ln k - k + \frac{1}{6} \ln (k + 4k^2 +8k^3) + \frac{1}{2} \ln \pi \right) \nonumber 
\end{eqnarray}
\begin{eqnarray}
\phantom{\log_2 \frac{n!}{k!(n-)!}}  
                &=& n \log_2 \frac{n}{n-k} + k \log_2 \frac{n-k}{k} + \\ 
                && + \frac{1}{6} \log_2 \frac{n+4n^2+8n^3}{((n-k)+4(n-k)^2+8(n-k)^3)(k+4k^2+8k^3)} - \frac{1}{2} \log_2 \pi \nonumber
\end{eqnarray}

We will take advantage of this expression in this note.

\end{document}